\documentclass[10pt,twocolumn,letterpaper]{article}
\usepackage{colortbl}
\usepackage[pagenumbers]{cvpr}
\usepackage{microtype}
\usepackage{multirow}
\usepackage{mathtools}
\usepackage{algorithm}
\usepackage{algorithmic}
\usepackage{enumitem}
\usepackage{balance}
\definecolor{cvprblue}{rgb}{0.21,0.49,0.74}
\usepackage[pagebackref,breaklinks,colorlinks,allcolors=cvprblue]{hyperref}
\newcommand{\method}{GEMS}

\DeclareMathOperator*{\argmin}{arg\,min}

\def\confName{CVPR}
\def\confYear{2027}

\title{\method: Generation and Experience Memory for
Subgoal-Guided Long-Horizon World Model Planning}
\author{
Yangchen Zhao$^{1}$ \quad
Ali Payani$^{2}$ \quad
Caiwen Ding$^{1}$\\
$^{1}$University of Minnesota, Twin Cities \quad
$^{2}$Cisco
}

\begin{document}
\maketitle

\begin{abstract}
Visual world models enable effective short-horizon goal-conditioned planning, such as reaching nearby targets and pushing objects over short distances, by predicting candidate action outcomes and comparing their representations with a goal representation. However, extending this approach to long-horizon tasks is challenging: prediction errors accumulate over extended rollouts, while latent-space similarity to distant goals provides limited guidance for ranking candidate outcomes. Existing methods address these limitations by introducing intermediate subgoals, either generated by learned models or selected from experience memory. Generated subgoals can contain hallucinations or be inaccurate in aspects such as object positions and timing. Memory-based approaches are constrained by experience coverage and may lack precise guidance for selecting appropriate intermediate states from stored observations. We introduce GEMS, which generates temporally structured subgoals to guide action search and state selection from Latent Experience Memory (LEM). A distance gate determines whether predicted outcomes are evaluated against the generated subgoal or the selected memory state. Experiments across six tasks and task suites spanning manipulation, navigation, and reaching demonstrate the effectiveness of GEMS for long-horizon planning. GEMS outperforms state-of-the-art world-model planning baselines across different manipulation settings, achieving success rates of 92.8\% on Cube, 68.4\% on Push-T, and 54.1\% on Meta-World. It improves over the strongest evaluated baseline on these tasks by 24.0, 7.3, and 22.2 percentage points, respectively. These results demonstrate that generated subgoals and experience memory can jointly strengthen long-horizon planning with a frozen world model.

\end{abstract}

\section{Introduction}

\begin{figure*}[t]
  \centering
  \includegraphics[width=\textwidth]{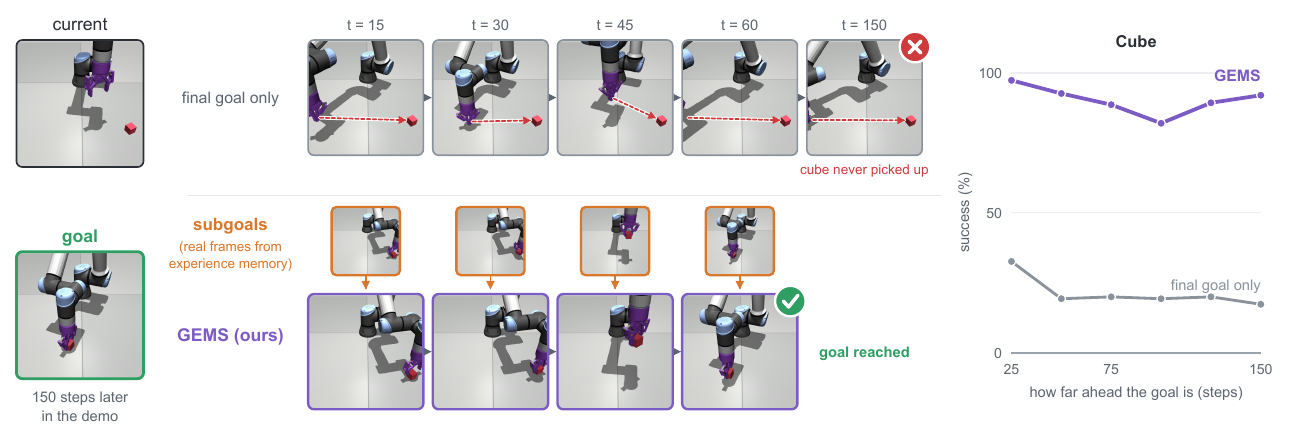}
    \caption{\textbf{Long-horizon planning with experience-grounded subgoals.}
    Left: planning toward the final goal alone never picks up the cube, whereas GEMS reaches the goal by planning toward intermediate subgoals. The subgoals (orange borders) are real frames selected from experience memory. Right: Cube success rates of GEMS and final-goal planning as the goal lies further ahead.}
    
  \label{fig:teaser}
\end{figure*}

Visual world models enable planning by predicting the consequences
of candidate actions before execution. Recent models make this
practical by learning action-conditioned dynamics in compact visual
representation spaces~\cite{dinowm,lewm,vjepa2}. In goal-conditioned planning, a common strategy is to search for
action sequences whose predicted outcomes are closest to the goal
representation.
However, prediction errors accumulate over long rollouts, while
latent-space similarity to distant goals becomes less informative
for distinguishing candidate outcomes~\cite{metro}. We therefore
study intermediate subgoals that guide a world-model planner
through shorter stages of a long-horizon task.

Prior work has explored both generated and experience-based
subgoals. Generative approaches predict intermediate images or
latent states to guide local control~\cite{hvf,pilot,hjepa}.
They offer flexible proposals, but generated targets can contain
hallucinations or be inaccurate in aspects such as object
positions and timing~\cite{metro}. Experience-based approaches, including Metro-WM~\cite{metro} and NGC~\cite{ngc}, use previously observed states to guide planning. Their guidance is grounded in
observed configurations, but it is limited by experience coverage.
Moreover, the current observation and the goal describe only the
present and the end of the task, so they give little guidance on
which observed state to target next. These approaches thus provide
complementary ingredients for subgoal construction.
Generation can propose an intermediate target for any state and
goal, while experience supplies observed configurations that can
serve as concrete planning references.

We introduce GEMS, Generation and Experience Memory for Subgoal
Planning. Its generator separately encodes the steps left to the
goal and the steps to the next subgoal, and predicts a learned
residual around a temporally weighted interpolation between
the current and final-goal representations. The resulting
proposal guides action candidates and state selection from
Latent Experience Memory (LEM), a bank of encoded observations
from training trajectories. A distance gate determines whether
predicted outcomes are evaluated against the generated subgoal
or the selected memory state. The encoder and
world model remain frozen during subgoal training
and planning. Fig.~\ref{fig:teaser} shows an example on Cube.

Our contributions are threefold:
\begin{enumerate}
    \item \textbf{Temporally structured subgoal generation.}
    We introduce a temporally structured subgoal formulation in which time shapes the prediction itself rather than serving only as an input. The remaining horizon and the subgoal offset are embedded separately, and their ratio places an interpolated anchor around which the subgoal is predicted as a residual.

    \item \textbf{Generation-guided experience memory.}
    We couple parametric subgoal generation with non-parametric
    Latent Experience Memory (LEM). Generated proposals guide
    the selection of stored observation latents, providing
    experienced references for intermediate planning targets.

    \item \textbf{Selective memory grounding.}
    We formulate a gated subgoal operator that selects between
    generated proposals and their matched experience states.
    Proposal-to-memory distance determines when to ground
    planning targets in experience and when to retain
    generated guidance.
\end{enumerate}

We evaluate GEMS on Cube and Cube-double~\cite{ogbench}, Push-T~\cite{dinowm}, TwoRoom and Reacher~\cite{lewm}, and Meta-World~\cite{metaworld}.
It achieves the best success on all four manipulation tasks, exceeding the strongest baseline by 24.0, 7.3, and 22.2 percentage points on Cube, Push-T, and Meta-World.
Ablations show that temporally structured generation and experience memory are both effective, and that keeping generated targets matters where memory coverage is sparse.

\section{Related Work}

\paragraph{Subgoal generation and hierarchical planning.}
Hierarchical planning and control have long used intermediate goals to decompose complex tasks into shorter-horizon problems~\cite{hpn,hiro}. Visual subgoal methods extend this principle to images and learned representations, allowing intermediate targets to be specified without explicit object-state coordinates. LEAP~\cite{leap} optimizes latent subgoal sequences using goal-conditioned value functions, while Hierarchical Foresight~\cite{hvf} searches for visual subgoals using a learned predictive model. More recently, PILOT~\cite{pilot} studies visual subgoals and policy priors for robot manipulation. Predictive hierarchies offer another way to organize planning across temporal scales: H-JEPA~\cite{hjepa} learns hierarchical world models, while Hi-LeWM~\cite{hilewm} examines hierarchical planning with LeWM~\cite{lewm}. These works establish intermediate visual targets as an interface between task-level guidance and local action selection. Closest to our work, SAGE~\cite{sage} conditions a latent subgoal generator on the remaining time and the subgoal offset and pairs it with an action prior for CEM planning with a frozen world model. GEMS keeps this generator--prior structure but changes how time enters the generator, through decoupled time embeddings and a progress-interpolated anchor, and selectively grounds the generated subgoals in experience.

\paragraph{Memory-based planning.}
Reusing stored experience is an established approach to planning in robotics and reinforcement learning.
Lightning~\cite{lightning} reuses and repairs previously solved motion paths, while E-Graphs~\cite{egraphs} organize past solutions into graphs that guide search.
In visual navigation, SPTM~\cite{sptm} and SoRB~\cite{sorb} connect observed states into planning graphs, while Sparse Graphical Memory~\cite{sgm} maintains a compact graph for efficient and robust planning.
ViNT~\cite{vint} combines generated visual subgoals with topological memory.
For world-model planning, Metro-WM~\cite{metro} builds graphs of experienced latents, NGC~\cite{ngc} scores candidates using graph-based geodesic distances, and Anchored Planning~\cite{anchored} uses experienced observations as intermediate targets.
IMWM~\cite{imwm} initializes CEM from similar demonstrations and uses reliability gating.
GEMS accesses memory through a time-aware generated proposal that selects an experienced scoring target from LEM; a distance gate retains the proposal when no sufficiently close match exists.

\section{\method}
\label{sec:method}

\begin{figure*}[t]
\centering
\includegraphics[width=\textwidth]{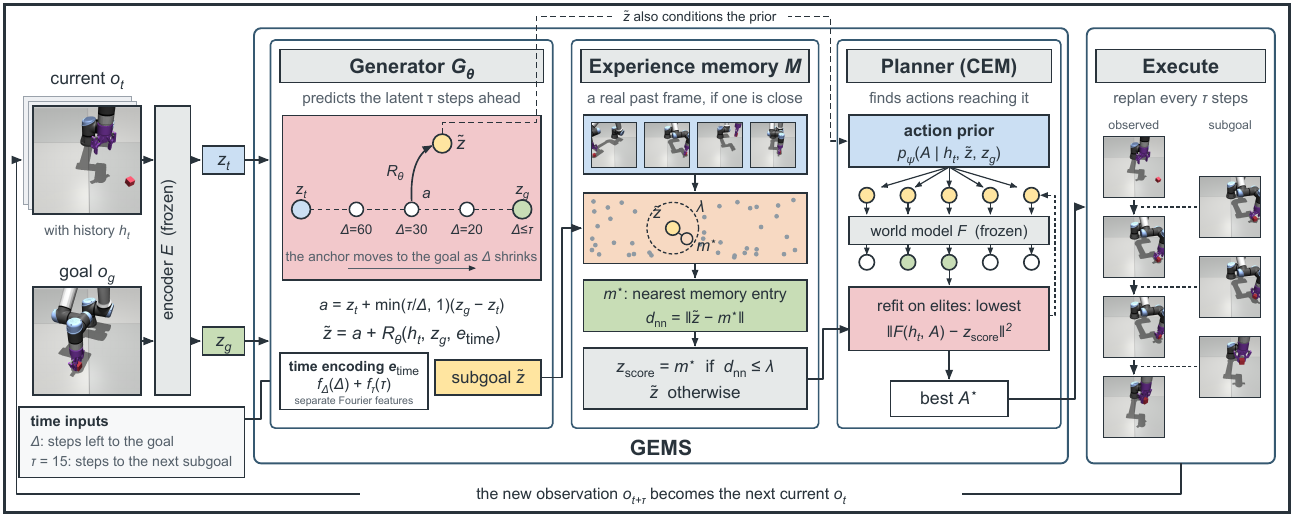}
\caption{\textbf{Overview of \method.} The generator proposes a subgoal $\tau$ steps ahead, and the experience memory grounds it in a stored frame when one lies close. CEM with the frozen world model plans $\tau$ actions toward the resulting target; \method\ executes them and replans. Right: one Cube rollout, with the frames observed after each segment and the subgoals they aimed at. For visibility, frames are re-rendered from the recorded states with a lighter floor.}

\label{fig:method_ge}
\end{figure*}

At each planning step, \method\ constructs a subgoal $\tau$
steps ahead and plans toward it with a frozen latent world
model $F$ and the cross-entropy method (CEM)~\cite{cem},
as shown in Figure~\ref{fig:method_ge}.
Three components construct the subgoal: temporally structured subgoal generation (Sec.~\ref{sec:method-generator}), generation-guided experience memory (Sec.~\ref{sec:lem}), and selective memory grounding (Sec.~\ref{sec:method-gate}).
Only the generator and the action prior are trained; the encoder and the world model stay frozen, and the experience memory is built once without training.
Section~\ref{sec:method-planning} describes planning and execution, and Algorithm~\ref{alg:method} summarizes the full procedure.

\begin{algorithm}[!t]
\caption{\method: time-aware, experience-grounded receding-horizon planning}
\label{alg:method}
\begin{algorithmic}[1]
\REQUIRE frozen encoder $E$ and world model $F$; generator $G_\theta$; action prior $p_\psi$; trajectories $\mathcal D_{\rm train}$, $\mathcal D_{\rm val}$; goal image $o_g$; goal horizon $H$; segment lengths $\tau_t$; budget $B\ge H$; CEM with $N$ candidates
\STATE $M\gets\{E(o)\,:\,o\in\mathcal D_{\rm train}\}$ \hfill $\triangleright$ Eq.~\ref{eq:lem}
\STATE $\lambda\gets$ best of $\{0,\lambda_{\rm nn},\lambda_e,\infty\}$ on $\mathcal D_{\rm val}$ \hfill $\triangleright$ Sec.~\ref{sec:method-gate}
\STATE $z_g\gets E(o_g)$;\; $t\gets0$
\WHILE{$t<B$ \AND goal not reached}
  \STATE $z_t\gets E(o_t)$;\; $\Delta_t\gets\max(H-t,\,\tau_t)$
  \IF{$\Delta_t>\tau_t$}
    \STATE $e_{\rm time}\gets W_o\operatorname{SiLU}\big(f_\Delta(\Delta_t)+f_\tau(\tau_t)\big)$ \hfill $\triangleright$ Eq.~\ref{eq:time}
    \STATE $a\gets z_t+\min(\tau_t/\Delta_t,1)\,(z_g-z_t)$ \hfill $\triangleright$ Eq.~\ref{eq:anchor}
    \STATE $\tilde z\gets a+R_\theta(h_t,z_g,e_{\rm time})$ \hfill $\triangleright$ Eq.~\ref{eq:gen}
    \STATE $m^\star\gets\argmin_{m\in M}\|\tilde z-m\|$ \hfill $\triangleright$ Eq.~\ref{eq:lookup}
    \STATE $z_{\rm score}\gets m^\star$ \textbf{if} $\|\tilde z-m^\star\|\le\lambda$ \textbf{else} $\tilde z$ \hfill $\triangleright$ Eq.~\ref{eq:projection}
  \ELSE
    \STATE $\tilde z\gets z_g$;\; $z_{\rm score}\gets z_g$ \hfill $\triangleright$ final segments
  \ENDIF
  \STATE $\{A^{(n)}\}_{n=1}^{N}\sim p_\psi(\cdot\mid h_t,\tilde z,z_g)$ \hfill $\triangleright$ initial candidates
  \STATE $A^\star\gets\operatorname{CEM}\big(\{A^{(n)}\};\,\|F(h_t,\cdot)-z_{\rm score}\|^2\big)$ \hfill $\triangleright$ Eq.~\ref{eq:cost}
  \STATE execute the $\tau_t$ actions of $A^\star$; observe $o_{t+\tau_t}$
  \STATE $t\gets t+\tau_t$;\; update $h_t$
\ENDWHILE
\end{algorithmic}
\end{algorithm}

\subsection{Temporally structured subgoal generation}
\label{sec:method-generator}
The generator $G_\theta$ predicts the latent of the state $\tau$ steps ahead.
The frozen encoder $E$ of the world model maps the current observation $o_t$ and the goal image $o_g$ to latents $z_t=E(o_t)$ and $z_g=E(o_g)$.
Besides $z_g$, the generator reads the history $h_t$, which holds the latents of the last three observations and, when the task provides it, the robot's proprioceptive state.
It also receives two time inputs: $\Delta$, the steps left to the goal, and $\tau$, the steps to the next subgoal.
From these inputs it builds the subgoal in three steps: it embeds the two time inputs, places an anchor between $z_t$ and $z_g$, and predicts a residual on top of the anchor.

\paragraph{Two clocks.}
The two time inputs differ in range: $\Delta$ can span the whole task, whereas $\tau$ only reaches the next subgoal.
We scale both by $D$, the largest goal offset seen in training.
A natural way to feed them to a network is to stack $\Delta/D$, $\tau/D$, and their ratio $\tau/\Delta$ into one vector and normalize it, which puts the three quantities on a common scale.
Normalizing them together, however, discards their overall scale: at $\Delta=D$, the vector $(1,\tau/D,\tau/D)$ normalizes to the same features for every $\tau$, so the generator can no longer tell how far ahead to predict.
\method\ therefore embeds the two time inputs separately.
With multi-frequency features
\begin{equation}
 \phi(x)=\big[x,\{\sin(\pi kx),\cos(\pi kx)\}_{k\in\{1,2,4\}}\big],
\end{equation}
the time embedding is
\begin{equation}
 e_{\rm time}=W_o\operatorname{SiLU}\!\big(f_\Delta(\Delta)+f_\tau(\tau)\big),
 \label{eq:time}
\end{equation}
where $f_\Delta(\Delta)=W_\Delta\,\phi(\Delta/D)$ and $f_\tau(\tau)=W_\tau\,\phi(\tau/D)$ are separate projections.
With its own projection and no normalization across inputs, each time input stays distinguishable over its whole range, so the generator can tell how much of the task remains apart from how much of it to complete now.

\paragraph{Residual prediction.}
The generator does not predict the subgoal latent from scratch.
It starts from an anchor $a$, a latent placed near where
the subgoal should be, and a Transformer
decoder~\cite{transformer} $R_\theta$ predicts only the
residual from the anchor to the subgoal:
\begin{equation}
 \tilde z=G_\theta(h_t,z_g,\Delta,\tau)=a+R_\theta(h_t,z_g,e_{\rm time}).
 \label{eq:gen}
\end{equation}
$R_\theta$ takes the history $h_t$, the goal latent $z_g$, and the time embedding $e_{\rm time}$ defined above.
With the anchor in place, $R_\theta$ only has to model how the subgoal differs from it rather than reproduce the whole scene.
What remains is where to place the anchor.

\paragraph{Progress-interpolated anchor.}
Two places suggest themselves.
When the goal is far, the next subgoal is only $\tau$ of the $\Delta$ remaining steps away, so the current latent $z_t$ is the natural anchor.
Near the deadline, the subgoal must reach the goal, so $z_g$ is.
\method\ moves the anchor between the two with the time inputs:
\begin{equation}
 a=z_t+w\,(z_g-z_t),\qquad w=\min(\tau/\Delta,\,1).
 \label{eq:anchor}
\end{equation}
The weight $w$ is the fraction of the remaining time that the next $\tau$ steps cover.
With $\tau=15$, the anchor lies a quarter of the way to the goal when $\Delta=60$, halfway when $\Delta=30$, and on the goal once $\Delta\le\tau$ (Figure~\ref{fig:method_ge}).
The true future latent need not lie on the line between $z_t$ and $z_g$; the residual corrects for that.
The time inputs therefore act twice: through $e_{\rm time}$, they tell $R_\theta$ how much time remains and how far ahead to predict, and through $w$, they set the starting point of the prediction.

\paragraph{Training.}
The generator is trained by latent regression on the training trajectories: for a step $t$, a goal $\Delta$ steps later, and an offset $\tau\le\Delta$, its prediction $\tilde z$ is matched to the encoded frame $z_{t+\tau}=E(o_{t+\tau})$.

\subsection{Generation-guided experience memory}
\label{sec:lem}
\paragraph{Experience memory.}
The latent experience memory (LEM) $M$ is a non-parametric record of every state visited in the training trajectories $\mathcal D_{\rm train}$, encoded by the same frozen encoder:
\begin{equation}
 M=\{E(o)\,:\,o\in\mathcal D_{\rm train}\}.
 \label{eq:lem}
\end{equation}
Each entry $m\in M$ is a real state expressed in the world model's own latent space, so it can serve directly as a planning target.
The memory is built once and never sees test episodes or object-state labels.

\paragraph{Generation as the key.}
A generated subgoal is time-aware and available for any state, but it is a regression estimate: it averages over plausible futures and can drift off the manifold of real states, for example by blurring the position of an object.
The memory contains only real states, but on its own it has no good key, since the current frame and the goal describe the present and the end rather than the next subgoal.
\method\ therefore lets the generator guide the memory: the proposal $\tilde z$ serves as the query, and its projection onto the memory under the Euclidean norm is
\begin{equation}
 m^\star=\argmin_{m\in M}\|\tilde z-m\|,\qquad
 d_{\rm nn}=\|\tilde z-m^\star\| .
 \label{eq:lookup}
\end{equation}
The projection $m^\star$ is the real state closest to the proposal, and $d_{\rm nn}$ measures how far the proposal must move to reach it.

\subsection{Selective memory grounding}
\label{sec:method-gate}
\label{sec:method-complement}
\paragraph{Distance gate.}
The projection helps only when a real state lies near the generated proposal.
Otherwise, the nearest stored state may show a different situation and mislead the planner.
\method\ therefore uses the projection only when $d_{\rm nn}$ is below a threshold $\lambda$:
\begin{equation}
 z_{\rm score}=P_{M,\lambda}(\tilde z)=
 \begin{cases}
 m^\star,& d_{\rm nn}\le\lambda,\\
 \tilde z,& \text{otherwise.}
 \end{cases}
 \label{eq:projection}
\end{equation}
The threshold is set per task on validation episodes.
We consider four values: $\lambda=0$, which never uses the memory; $\lambda=\infty$, which always uses it; and two values measured on validation data, the median distance $\lambda_{\rm nn}$ between generated subgoals and their nearest stored states, and the median error $\lambda_e$ of the generator.
\method\ uses the value with the most validation successes.

\paragraph{Complementarity.}
The complete subgoal operator composes generation and grounding,
\begin{equation}
 Q_{\theta,\lambda}(h_t,z_g,\Delta,\tau)=P_{M,\lambda}\big(G_\theta(h_t,z_g,\Delta,\tau)\big).
 \label{eq:unified}
\end{equation}
Generation supplies a time-aware proposal and a query at each planning step, and the memory supplies a real state wherever the training data cover the proposal.
Validation sets how far to trust the memory on each task (Sec.~\ref{sec:memory-complement}).

\subsection{Planning and execution}
\label{sec:method-planning}
At each planning step, the planner turns the subgoal into actions with the frozen world model $F$.
Given the history $h_t$ and a candidate action sequence $A$ for the next $\tau_t$ steps, $F$ predicts the latent the scene will reach, and the planner seeks the actions whose prediction is closest to the scoring target:
\begin{equation}
 J(A;z_{\rm score})=\big\|F(h_t,A)-z_{\rm score}\big\|^2 .
 \label{eq:cost}
\end{equation}
CEM minimizes $J$, starting from candidates proposed by an action prior $p_\psi(A\mid h_t,\tilde z,z_g)$ and returning the final elite mean as the plan $A^\star$.
The prior is trained with the generator's own proposals as its condition, so it reads $\tilde z$ even when the cost uses a stored state.

The agent executes the $\tau_t$ actions of $A^\star$, observes the new state, and replans with a new subgoal.
With goal horizon $H$, the time left at step $t$ is $\Delta_t=\max(H-t,\tau_t)$, where $\tau_t$ is 15 steps by default.
Once $\Delta_t\le\tau_t$, the agent plans toward $z_g$ directly until the goal is reached or the episode budget $B\ge H$ runs out.

\section{Experiments}

\paragraph{Tasks and evaluation.}
We evaluate on six tasks: Cube and Cube-double from OGBench~\cite{ogbench}, Push-T~\cite{dinowm}, TwoRoom and Reacher~\cite{lewm}, and Meta-World~\cite{metaworld}, a suite of six manipulation tasks.
The goal image is the frame $H$ steps after the start state in a held-out test trajectory, with $H=150$ on Cube, Push-T, and Reacher, $H=100$ on TwoRoom, and $H=75$--$150$ on Meta-World. Cube-double uses OGBench's official goals, which take 75--255 steps.
An episode succeeds if the task's success test is met within the step budget: 150 steps on Cube, 500 on Cube-double, and $2H$ on the other tasks.
We use 400, 250, 150, 150, 150, and 300 test queries, respectively.
Each query is run with three planner seeds, except on TwoRoom and Reacher.
We report the success rate over all test episodes.

\paragraph{Data and models.}
Each task's data are split by episode into training, validation, and test sets.
Our generator, action prior, and experience memory are built from the training split, and validation data are used only to select settings.
All methods except H-JEPA share the same frozen LeWM~\cite{lewm} for each task.
We use the released checkpoints where available and train LeWM on our training split otherwise.
No method observes object states.
\method\ predicts subgoals $\tau=15$ steps ahead and plans with CEM, replanning every $\tau$ steps.
Following SAGE~\cite{sage}, CEM uses 300 candidates, 30 iterations, and 30 elites, initialized by an action prior.
The experience memory of \method\ stores the encoded frames of the training trajectories, and the memory option is selected per task on validation data.

\paragraph{Baselines.}
We compare with eight planners. Flat LeWM~\cite{lewm} plans toward the final goal without subgoals, and SAGE~\cite{sage} generates latent subgoals with a subgoal-conditioned action prior.
Metro-WM~\cite{metro} and NGC~\cite{ngc} use graphs of training states to choose waypoints or to measure cost, while PRISM~\cite{prism} and LeFlow~\cite{leflow} learn to propose actions or latent plans.
Hi-LeWM-C~\cite{hilewm} and H-JEPA~\cite{hjepa} plan hierarchically.
All baselines use our queries, step budgets, and success tests, and their learned components are trained on our training split with the authors' recipes.
The exception is H-JEPA, for which we use the released weights.
These exist only for Cube and PushT and were trained on episodes that include our test episodes (marked \textsuperscript{\dag}).

\subsection{Comparison with Existing Methods}
\label{sec:main-results}

Table~\ref{tab:external} reports closed-loop success on the six tasks.
\method\ achieves the highest success on all four manipulation tasks: 92.8\% on Cube, 68.4\% on PushT, 17.7\% on Cube-double, and 54.1\% on Meta-World.
These are 24.0, 7.3, 16.9, and 22.2 percentage points above the best competing method on each task.
On Cube-double, all other methods stay below 1\%.
TwoRoom and Reacher are close to saturation for the stronger methods, and \method\ reaches 98.0\% and 94.7\%, within 1.3 points of the best method on each.
Averaged over the six tasks, \method\ reaches 71.0\%, compared with 59.2\% for the strongest baseline.

\begin{table*}[!t]
\centering
\small
\setlength{\tabcolsep}{7pt}
\caption{\textbf{Closed-loop success on six tasks (\%).} $H$ is the goal horizon in environment steps. $^\dagger$Released weights, trained on episodes that include our test episodes.}
\label{tab:external}
\begin{tabular}{lrrrrrr}
\toprule
Method & Cube & PushT & Cube-double & TwoRoom & Reacher & Meta-World\\
Horizon $H$ & 150 & 150 & 75--255 & 100 & 150 & 75--150\\
\midrule
Flat LeWM~\cite{lewm}
& 17.8 & 4.7 & 0.0 & 14.7 & 68.0 & 0.0\\
LeFlow~\cite{leflow}
& 29.8 & 12.7 & 0.0 & 86.0 & 82.7 & 0.6\\
Hi-LeWM-C~\cite{hilewm}
& 25.4 & 22.7 & 0.0 & 38.7 & 86.0 & 0.1\\
H-JEPA$^\dagger$~\cite{hjepa}
& 26.3 & 22.9 & -- & -- & -- & --\\
PRISM~\cite{prism}
& 36.6 & 16.0 & 0.1 & 63.3 & 79.3 & 0.2\\
NGC~\cite{ngc}
& 57.8 & 11.1 & 0.0 & 95.3 & \textbf{95.3} & 0.1\\
Metro-WM~\cite{metro}
& 59.7 & 37.8 & 0.0 & \textbf{99.3} & \textbf{95.3} & 2.8\\
SAGE~\cite{sage}
& 68.8 & 61.1 & 0.8 & 98.0 & 94.7 & 31.9\\
\midrule
\method
& \textbf{92.8} & \textbf{68.4} & \textbf{17.7} & 98.0 & 94.7 & \textbf{54.1}\\
\bottomrule
\end{tabular}
\end{table*}

\subsection{Planning across Goal Horizons}
\label{sec:task-difficulty}

We evaluate Cube and PushT at goal horizons $H\in\{25,50,75,100,125,150\}$ and give each episode $2H$ steps.
Each horizon from 25 to 125 uses 150 queries and one planner seed.
At $H=150$, we reuse all test episodes from Table~\ref{tab:external}, also with $2H$ steps.
\method\ uses the same configuration as in Table~\ref{tab:external}.

\begin{figure}[t]
\centering
\includegraphics[width=\linewidth]{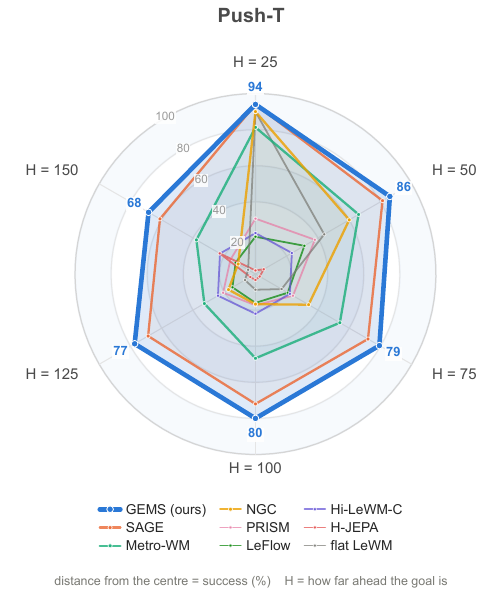}
\caption{\textbf{Planning across goal horizons on PushT.} Each axis is a goal horizon $H$; a larger radius means higher success.}
\label{fig:pusht-horizon}
\end{figure}

\paragraph{PushT.}
Figure~\ref{fig:pusht-horizon} shows that \method\ matches SAGE at $H=25$, where both reach 94.0\%, and outperforms all baselines at every longer horizon.
At $H=150$, \method\ reaches 68.4\%, compared with 61.1\% for SAGE, 37.8\% for Metro-WM, and 4.7\% for Flat LeWM.
\method\ thus keeps a clear lead as soon as the goal is more than a few dozen steps away.

\begin{figure}[t]
\centering
\includegraphics[width=\linewidth]{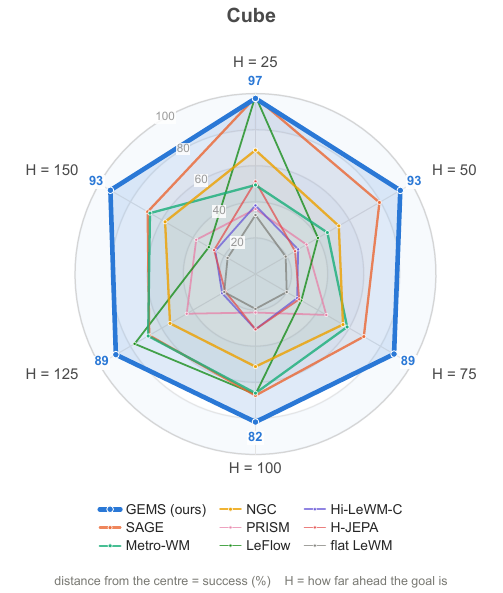}
\caption{\textbf{Planning across goal horizons on Cube.} Each axis is a goal horizon $H$; a larger radius means higher success.}
\label{fig:cube-difficulty}
\end{figure}

\paragraph{Cube.}
Figure~\ref{fig:cube-difficulty} shows that at $H=25$, \method\ (97.3\%) is close to SAGE and LeFlow (98.0\% and 98.7\%).
From $H=50$ onward, \method\ leads all baselines and stays above 80\%.
At $H=150$, it reaches 92.8\%, compared with 69.0\% for SAGE and 17.8\% for Flat LeWM.
Long horizons barely affect \method\ on Cube: its success drops by only 4.5 points from $H=25$ to $H=150$.

\paragraph{Near and far goals.}
Near goals do not separate the methods: at $H=25$, several baselines are as strong as \method\ on both tasks.
As $H$ grows, their success drops steeply: Flat LeWM falls from 90.0\% to 4.7\% on PushT, and LeFlow from 98.7\% to 29.8\% on Cube.
\method\ declines much less, so it is strong at every horizon, and its lead is largest when the goal is far.

\subsection{Gap to Expert Subgoals}
\label{sec:expert-gap}

We compare each method with an expert reference that plans toward the expert's own future frames.
At each planning step, the reference uses the frame that the expert reached 15 steps later as its target, instead of a generated subgoal.
We then measure how much of the gap between Flat LeWM and this reference each method closes:
\begin{equation}
R_m = 100\,\frac{S_m-S_{\mathrm{flat}}}{S_{\mathrm{expert}}-S_{\mathrm{flat}}},
\label{eq:expert-gap}
\end{equation}
where $S$ denotes success rate.
All methods use the same test episodes and planner seeds as the reference.
On Cube, the reference was run with one planner seed, so the rates there differ slightly from the three-seed averages in Table~\ref{tab:external}.

\begin{figure}[t]
\centering
\includegraphics[width=\linewidth]{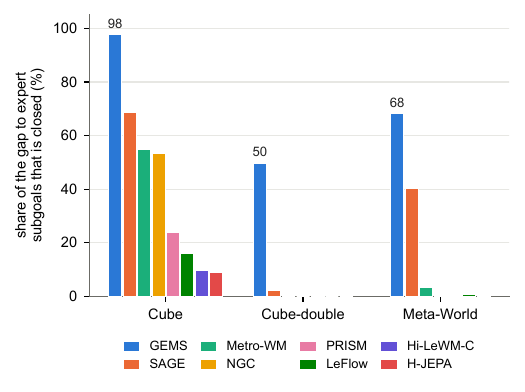}
\caption{\textbf{Closing the gap to expert subgoals.} Gap closed follows Eq.~\ref{eq:expert-gap}: 0\% is Flat LeWM and 100\% is the expert reference.}
\label{fig:expert-gap}
\end{figure}

The expert reference reaches 94.0\%, 35.6\%, and 79.3\% success on Cube, Cube-double, and Meta-World.
Figure~\ref{fig:expert-gap} shows that \method\ closes 97.7\%, 49.8\%, and 68.2\% of the gap, while the best baseline on each task closes 68.6\%, 2.2\%, and 40.2\%.
\method\ thus nearly matches planning with expert subgoals on Cube, and on the two harder tasks it closes far more of the gap than any baseline.

\subsection{Where Do the Gains Come From?}
\label{sec:gain-source}

\begin{figure}[t]
\centering
\includegraphics[width=\linewidth]{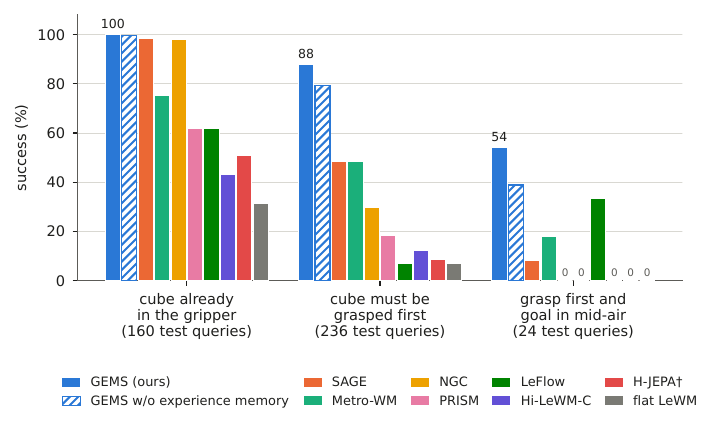}
\caption{\textbf{Cube success by start condition.} Success within 150 steps over three planner seeds. The mid-air group is a subset of the grasp-first group. Four queries already within the 4\,cm success radius are excluded}
\label{fig:grasp-start}
\end{figure}

We split the Cube test queries by whether the cube is already in the gripper at the start or must first be grasped (Figure~\ref{fig:grasp-start}).
When the cube is already held, \method\ and the best baseline both succeed almost always (100.0\% and 98.3\%).
When the cube must first be grasped, \method\ reaches 87.9\%, while no baseline exceeds 48.4\%.
The advantage of \method\ thus comes from the hardest part of the task: picking up the object before moving it.

Experience memory contributes in the same place.
Without it, success on grasp-first queries drops from 87.9\% to 79.4\%, while in-hand queries barely change (100.0\% vs.\ 99.6\%).
The hardest case is the 24 grasp-first queries whose goal holds the cube in mid-air.
Here \method\ reaches 54.2\%, compared with 38.9\% without memory and at most 33.3\% for any baseline.
Memory therefore helps most on exactly the queries that are hardest to plan.

We also examine what the memory changes in the subgoals.
We use a probe that reads the cube's position from a latent.
On validation segments where the cube is still on the table 15 steps later, the probe finds the cube already lifted in 15.5\% of generated subgoals: the generator skips ahead past the grasp.
After the memory replaces them with real frames, this drops to 3.2\%, close to the probe's own error of 1.3\% on real future frames.
The memory thus stops subgoals from skipping the grasp, which matches where its gains appear.

\subsection{Generation and Memory Are Complementary}
\label{sec:memory-complement}

As discussed earlier, when the memory covers the task densely, it replaces imprecise generated subgoals with nearby real frames; when it is sparse and lacks the states the task requires, the generator takes over.
We test both cases on Cube and Cube-double.

\begin{figure*}[!t]
\centering
\includegraphics[width=\textwidth]{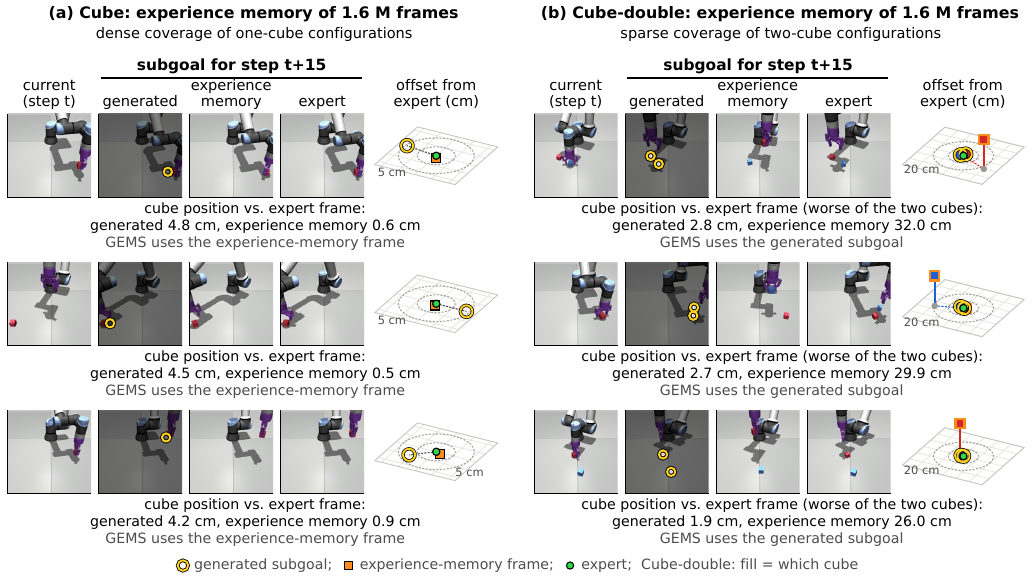}
\caption{\textbf{Generated subgoals, experience-memory frames, and expert frames.} Each row shows the current frame, three subgoals for 15 steps later, and the 3D offset of each subgoal's cube from the expert's. The generated column shows the dimmed expert frame with a yellow ring at the cube position read from the generated latent.}
\label{fig:memory-examples}
\end{figure*}

Figure~\ref{fig:memory-examples} shows both cases with memories of the same size.
On Cube, the memory frame places the cube within 1\,cm of the expert, while the generated subgoal is 4--5\,cm away.
On Cube-double, the memory frame misplaces one cube by 26--32\,cm, while the generated subgoal keeps both within 3\,cm.
Over all validation subgoals, the memory frame is closer to the expert in 69\% of Cube cases, and the generated subgoal is closer in 77\% of Cube-double cases.
To tell whether the memory is limited by the lookup or by the data, we also query it with the expert's true future frame, which returns the closest frame the memory holds.
On Cube, the frame selected by the generated subgoal is off by 0.9\,cm (median), close to this best case of 0.6\,cm.
On Cube-double, even the best case is off by 2.7\,cm for the worse cube, more than four times the Cube value, because two cubes span far more configurations than one.

\method\ therefore combines the two sources instead of choosing one.
It gets the best of both: the precision of real experience where the data are dense, and the reach of generation where they run out.

\subsection{Ablation Studies}
\label{sec:ablations}

We ablate the two parts of temporally structured subgoal generation, the interpolated anchor and the separate time encoding, together with the experience memory (Table~\ref{tab:components}).
Each variant has its own matched action prior and memory option.

\begin{table*}[!t]
\centering
\small
\caption{\textbf{Ablations (success, \%).} (a) Generator anchor. (b) Time encoding and experience memory.}
\label{tab:components}
\begin{minipage}[t]{.41\textwidth}
\centering
\textbf{(a) Generator anchor}\par\smallskip
\setlength{\tabcolsep}{3pt}
\begin{tabular}{lrrrr}
\toprule
Anchor & Cube & Push-T & TwoRoom & Reacher\\
\midrule
Final goal    & 88.8 & 67.3 & 98.7 & 95.3\\
Current state & 91.4 & 66.9 & 96.0 & 90.7\\
\rowcolor{gray!12}
Interpolation & 92.8 & 68.4 & 98.0 & 94.7\\
\bottomrule
\end{tabular}
\end{minipage}
\hfill
\begin{minipage}[t]{.56\textwidth}
\centering
\textbf{(b) Time encoding and experience memory}\par\smallskip
\setlength{\tabcolsep}{3pt}
\begin{tabular}{llrrrrr}
\toprule
Time enc. & Memory & Cube & Push-T & TwoRoom & Reacher & Meta-World\\
\midrule
joint    & no  & 67.6 & 63.1 & 97.3 & 90.7 & 41.6\\
separate & no  & 87.7 & 70.4 & 98.0 & 92.7 & 46.2\\
joint    & yes & 84.6 & 64.4 & 97.3 & 96.0 & 48.0\\
\rowcolor{gray!12}
separate & yes & 92.8 & 68.4 & 98.0 & 94.7 & 54.1\\
\bottomrule
\end{tabular}
\end{minipage}
\end{table*}

\paragraph{Interpolated anchor.}
Table~\ref{tab:components}(a) anchors the generator at the final goal, the current state, or their interpolation.
Interpolation is best on Cube and Push-T and within 0.7 points of the best on TwoRoom and Reacher, while the two fixed anchors fall up to 4.0 and 4.6 points behind.
Interpolation is therefore the most reliable anchor across tasks.

\paragraph{Time encoding and experience memory.}
Table~\ref{tab:components}(b) builds \method\ up from a variant that embeds the time inputs jointly (Sec.~\ref{sec:method-generator}) and uses no memory.
Separate encoding brings large gains: without memory, it adds 20.1, 7.3, and 4.6 points on Cube, Push-T, and Meta-World.
Experience memory then adds 5.1, 2.0, and 7.9 points on Cube, Reacher, and Meta-World.
Together, they improve on the variant with neither by more than 25 points on Cube and 12 points on Meta-World.
Both designs are therefore effective, and combining them yields strong gains.

\section{Conclusion}
We introduced \method, combining temporally structured
subgoal generation with selective experience memory for
long-horizon world-model planning.
Across six tasks and task suites, \method\ improves
manipulation performance while maintaining strong navigation
and reaching results.
Our analyses show how experience refines generated targets
and why generation remains useful when memory coverage
is limited.

\clearpage
\section*{LLM Usage Statement}
OpenAI Codex and Anthropic Claude were used to assist with research discussion, code development, and language refinement. The authors are responsible for the correctness, attribution, and final verification of the submission.

{\small
\bibliographystyle{ieeenat_fullname}
\bibliography{references}
}

\end{document}